%% file: main.tex
\documentclass[lettersize,journal]{IEEEtran}
\usepackage{amsmath,amsfonts}
\usepackage{array}
\usepackage[caption=false,font=normalsize,labelfont=sf,textfont=sf]{subfig}
\usepackage{textcomp}
\usepackage{stfloats}
\usepackage{url}
\usepackage{verbatim}
\usepackage{graphicx}
\usepackage{bbm}
\usepackage{makecell}
\usepackage{graphicx}
\usepackage{multirow}
\usepackage[table]{xcolor}
\usepackage{siunitx}
\usepackage{xspace}
\usepackage{bbold}
\usepackage{enumitem}
\usepackage{float}
\usepackage{wrapfig}
\usepackage{amsmath}
\usepackage{algorithm}
\usepackage[noEnd=true,indLines=true]{algpseudocodex}
\usepackage{booktabs}
\usepackage{etoc}
\etocdepthtag.toc{mtchapter}
\etocsettagdepth{mtchapter}{subsubsection}
\etocsettagdepth{mtappendix}{none}
\usepackage[numbers,sort&compress]{natbib}
\usepackage{hyperref}

\usepackage{amssymb}

\begin{document}

\title{Asking Less, Adapting Better: \\
       Feedback-Efficient Test-Time Adaptation for Vision-and-Language Navigation}

\author{Zixu Wang$^*$, Huaquan Zhang$^*$, Shoukai Xu$^*$, Jinwu Hu, Zhiquan Wen$^\dag$, Mingkui Tan, \IEEEmembership{Member IEEE}

\thanks{Zixu Wang is with the School of Software Engineering, South China University of Technology, Guangzhou, China, also with the Institute for Super Robotics (Huangpu), Guangzhou, China. (e-mail: sewangzixu@mail.scut.edu.cn).}

\thanks{Huaquan Zhang, Shoukai Xu, Jinwu Hu and Zhiquan Wen are with the School of Software Engineering, South China University of Technology, Guangzhou, China. (e-mail: sezhq@mail.scut.edu.cn, shoukaixv@gmail.com, fhujinwu@gmail.com, sewenzhiquan@gmail.com).}

\thanks{Mingkui Tan is with the School of Software Engineering, South China University of Technology, and also with the Key Laboratory of Big Data and Intelligent Robot, Ministry of Education, Guangzhou 510006, China (e-mail: mingkuitan@scut.edu.cn).}

\thanks{$^*$Authors contributed equally. $^\dag$Corresponding authors}

}

\markboth{Journal of \LaTeX\ Class Files,~Vol.~14, No.~8, August~2021}%
{Shell \MakeLowercase{\textit{et al.}}: A Sample Article Using IEEEtran.cls for IEEE Journals}


\newcommand{\Entropy}{H}
\newcommand{\methodname}{FETA\xspace}
\newcommand{\fullmethodname}{\textbf{F}eedback-\textbf{E}fficient \textbf{T}est-Time \textbf{A}daptation method\xspace}

\def\blue{\textcolor{black}}

\maketitle

\begin{abstract}
Vision-and-Language Navigation (VLN) requires agents to follow instructions and navigate to target locations. However, current VLN agents often suffer severe performance degradation in unseen environments due to distribution shifts. Test-Time Adaptation (TTA) offers a promising paradigm to bridge this gap, yet existing methods that rely on unsupervised entropy minimization can propagate errors through the trajectory, causing cascading failures in navigation. Recent methods mitigate this by incorporating human feedback, but they apply trajectory-level supervision uniformly across all steps without localizing the actions is truly informative, leading to inefficient feedback utilization and mismatched optimization granularity. To address these limitations, we propose \methodname{}, a \fullmethodname{} that improves both feedback efficiency and optimization reliability.
To reduce unnecessary human interventions, we introduce a hierarchical sample selection strategy to localize informative samples. Specifically, we first filter trajectories whose normalized policy entropy exhibits large fluctuations. Within each selected trajectory, we further select the critical action via an entropy-gradient saliency criterion and query a single binary feedback on its correctness.
A discrepancy-aware entropy optimization objective then updates the policy only at the queried step, with adaptive scaling that assigns larger relative weights to samples with greater prediction--feedback discrepancy. Extensive experiments across multiple benchmarks demonstrate that \methodname{} achieves state-of-the-art performance with fewer human interventions, while remaining robust under noisy feedback and continual adaptation settings.
\end{abstract}


\section{Introduction}
\IEEEPARstart{V}{ision}-and-Language Navigation (VLN)~\cite{anderson2018vision, qiao2022hop, chen2022think, hong2022brige} is a fundamental embodied AI task in which an agent interprets natural-language instructions and navigate through a 3D environment to reach a target location. In practical applications, VLN agents are often deployed under conditions that differ from those seen during training. Such deployment shifts may arise from new room layouts, varying illumination or different instruction granularity, leading to substantial discrepancies between the training and test distributions~\cite{guhur2021airbert, gao2024fastslow, ko2025active, lin2021adversarial, wang2023learning}. As a result, even strong VLN models often suffer marked performance drop when transferred to new environments, making online adaptation crucial for robust embodied navigation~\cite{an2024etpnav, wang2023scaling, an2023bevbert, Wang2024GOAT, wang2026lifelong}.

Test-Time Adaptation (TTA)~\cite{wang2021tent,niu2022efficient, niutest, su2024revisiting} offers a natural way to mitigate such distribution shifts by updating model parameters on incoming test data. Motivated by its success in static vision tasks such as image classification and semantic segmentation, recent studies have begun to extend TTA to VLN. However, existing methods~\cite{gao2024fastslow, hong2025general} typically rely on unsupervised objectives, such as entropy minimization, and perform policy updates at every step along the navigation trajectory. This strategy is particularly brittle in VLN, where navigation unfolds as a sequential decision-making process: an erroneous update at one step can change the subsequent state distribution, distort later optimization signals, and eventually cause cascading failures. Therefore, the key challenge in VLN test-time adaptation is not simply how to adapt online, but how to derive optimization signals that remain reliable throughout sequential decision-making.

To obtain more reliable optimization signals, recent VLN-TTA methods have begun to incorporate human feedback at test time. FeedTTA~\cite{kim2025test} queries human whether each trajectory is successful or not, and uses the feedback as a reward signal to update the policy via reinforcement learning. ATENA~\cite{ko2025active} requests feedback only for trajectories with high entropy, calibrating policy uncertainty by boosting decision confidence on successful trajectories and suppressing it on failed ones. Despite their promise, these methods still suffer from two key limitations. \textit{1) Inefficient Feedback Utilization}: Once a trajectory-level label (\textit{i.e.,} success or failure) is obtained, existing methods generally use every obtained label to update the policy on the corresponding trajectory, even when the resulting adaptation gains are limited. \textit{2) Mismatched Optimization Granularity}: Existing methods use the same trajectory-level feedback signal across all steps, even though successful trajectories may still contain suboptimal actions and failed ones may still include locally correct decisions. This creates a mismatch between trajectory-level supervision and local action decisions being optimized.

To address these challenges, we propose a \fullmethodname{} (\methodname{}) that improves both feedback efficiency and optimization reliability. Our key observation is that the choice of trajectories and steps for feedback acquisition affects adaptation performance (as shown in Fig.~\ref{fig:trajectory_selection_ablation} and Table~\ref{tab:ablation}). Based on this observation, we design a hierarchical feedback selection strategy that improves both feedback efficiency and optimization granularity. At the trajectory level, we prioritize trajectories with large fluctuations in normalized policy entropy, as such trajectories indicate unstable decision confidence and may benefit from external correction. At the step level, for each selected trajectory, we select the critical decision state using an entropy-gradient saliency criterion, which measures the sensitivity of normalized policy entropy to the action logits. We then query a single binary human feedback signal on the correctness of the selected action. This hierarchical design significantly reduces unnecessary human intervention while refining supervision from the trajectory level to the action level.

Based on the queried feedback, we design a discrepancy-aware objective for policy optimization. Instead of updating the entire trajectory from a single final outcome, our method corrects the policy only at the queried step, where the feedback determines both the update magnitude and direction of entropy optimization. Specifically, we introduce an adaptive scaling strategy that assigns larger relative weights to samples with greater mismatch between the current policy prediction and the feedback. We then apply directional entropy optimization: for actions verified as correct, the objective sharpens the policy distribution to reinforce the current decision. Otherwise, it relaxes overconfident predictions and encourages alternative actions. Our contributions can be summarized as follows:
\begin{itemize}[leftmargin=1.5em, labelsep=0.5em]
    \item We propose a \fullmethodname{} for VLN and substantially reduce the number of human interventions through a hierarchical sample selection strategy, achieving a favorable trade-off between feedback cost and performance gain. 
    \item We design a discrepancy-aware entropy optimization loss that dynamically adjusts the direction of entropy optimization and the relative sample weights according to the human feedback, enabling efficient utilization of sparse feedback signals. 
    \item Extensive experiments show that \methodname{} achieves the state-of-the-art performance across multiple benchmarks with fewer human interventions, including an 18.30\% relative improvement in SPL over the strongest existing navigation TTA baseline, using DUET as the backbone on the REVERIE val unseen split, while remaining robust under noisy feedback and continual adaptation.
\end{itemize}

\section{Related Work}
\subsection{Vision-and-language navigation}
Vision-and-language navigation (VLN)~\cite{anderson2018vision, hong2022brige, shi2025smartway, lin2026evolvenav} is a fundamental task in embodied AI, requiring an agent to translate natural-language instructions into grounded perception and sequential actions~\cite{he2026fine}. Early VLN methods were largely based on recurrent seq2seq architectures, augmented with mechanisms such as speaker-based data augmentation, progress estimation, and search to improve instruction following and long-horizon decision making~\cite{NEURIPS2018_speaker, ma2019selfmonitoring, Wang2020ActiveVI, wang2020vision}. Subsequent work shifted toward pretraining and Transformer-based models~\cite{guhur2021airbert, wang2023scaling, Hong_2021_CVPR, chen2021hamt, zhang2024uni, wei2025streamvln, qiao2023hop+} which substantially improved multimodal fusion, history modeling, and benchmark performance. Despite this progress, most VLN models are still trained offline and tend to suffer significant performance degradation when deployed in unseen environments~\cite{lin2024correctable}. Recent efforts explore zero-shot navigation with Large Language Models (LLMs)~\cite{qiao2025opennav, zhou2024navgpt, chen2023a2navactionawarezeroshotrobot, long2024instructnav, qiaonavbench, chen2025constraint} to improve generalization without task-specific training. However, their reliance on large models incurs high inference latency, limiting practicality for real-time navigation. Another line of work studies test-time adaptation through online updates at deployment. Existing methods typically optimize self-supervised objectives such as entropy minimization~\cite{gao2024fastslow, wang2021tent, niu2022efficient, zhang2025come}, which are weakly aligned with navigation objectives and can lead to brittle adaptation under distribution shifts. In contrast, our work leverages binary human feedback to directly align adaptation with the navigation goal, enabling more stable updates in unseen environments. 

\subsection{Test-Time Adaptation}
Test-Time Adaptation (TTA) aims to update the model parameters directly using target domain data during inference, without access to source domain labels~\cite{dong2026advances}. Most methods~\cite{wang2021tent, niu2022efficient, wang2022continual, niu2023towards, tan2025uncertainty} adopt entropy minimization as the optimization objective, improving the confidence of policy predictions to enable online adaptation to unseen domains, and have demonstrated strong performance on tasks such as image classification~\cite{lee2025testtime, karmanov2024efficient} and semantic segmentation~\cite{cao2023multi, lee2023towards}. Recent work has begun to explore the application of TTA to Vision-and-Language Navigation. FSTTA~\cite{gao2024fastslow} achieves a balance between fast adaptation and long-term stability through gradient and parameter decomposition. FeedTTA~\cite{kim2025test} incorporates binary trajectory-level human feedback as a reinforcement learning reward signal to update the navigation policy online. ATENA~\cite{ko2025active} selectively requests feedback only for high-entropy trajectories and calibrates policy confidence accordingly. However, existing methods that rely on human feedback use the same trajectory-level feedback signal across all steps, potentially reinforcing suboptimal actions or suppressing locally correct ones. By contrast, our method uses hierarchical feedback selection and improving adaptation performance while requiring fewer feedback queries.

\section{Problem Statement and Motivation}
Vision-and-Language Navigation (VLN) is a partially observable sequential decision-making task, where an agent follows the instruction to navigate in a 3D environment. Given an instruction $I$, at each time step $t$ the agent receives a visual observation $o_t$ and maintains a history representation $h_t$, forming the agent state $s_t = (o_t, h_t, I)$. The agent then selects the highest-probability action $a_t$ from the candidate set $\mathcal{A}_t$ according to its policy $\pi_{\theta}$. The navigation continues until the agent predicts \texttt{STOP} or reaches the maximum step limit, yielding a trajectory $\tau=\{(o_t,a_t)\}_{t=1}^{T}$.

In real deployment, VLN agents often encounter environments that differ from those seen during training, which can degrade navigation performance. This motivates test-time adaptation (TTA), which updates model parameters online using incoming test samples. Existing TTA methods typically rely on unsupervised surrogate objectives, \textit{e.g.,} entropy minimization. However, unlike static perception tasks, VLN is inherently sequential: each action affects future states, observations, and subsequent optimization signals. As a result, once the adaptation signal is unreliable, an erroneous update can propagate through the trajectory and cause cascading failures. The challenge is how to obtain reliable optimization signals under sequential decision-making and environment shift.

To improve adaptation reliability, recent VLN test-time adaptation methods have begun to introduce reliable human feedback during deployment. However, existing human-feedback-driven methods still remain limited in two important aspects. First, the feedback they acquire may not always translate into effective adaptation, as policy updates on some labeled trajectories yield only limited gains. Second, they typically use a single trajectory-level feedback signal to optimize all actions within that trajectory in the same direction. Such coarse supervision is mismatched to the fine-grained nature of sequential decision-making: successful trajectories may still contain suboptimal actions, while failed trajectories may still include locally correct decisions. Applying the same feedback to all steps can therefore provide inappropriate optimization signals for individual actions, reducing the effectiveness of human guidance for policy adaptation. These limitations suggest that human-feedback-driven TTA for VLN requires a more structured use of sparse supervision, with both feedback-efficient querying and fine-grained action-level correction.

\begin{algorithm}[t]
    \caption{The pipeline of \fullmethodname for VLN}
    \label{alg:workflow}
    \begin{algorithmic}[1]
        \Require Pretrained policy $\pi_\theta$, threshold $\delta_\sigma$,
        \Statex scaling exponent $\gamma$, learning rate $\eta$
        \For{each episode}
            \State Collect trajectory $\tau = \{(o_t, a_t)\}_{t=1}^{T}$ using $\pi_\theta$
            \State Compute $\sigma_{\hat{H}}(\tau)$ via Eq.~(\ref{eq:traj_std})
            \If{$\sigma_{\hat{H}}(\tau) > \delta_\sigma$}
                \State Compute $G_t$ via Eq.~(\ref{eq:saliency})
                \State Select step $t^* = \arg\max_t \, G_t$
                \State Query binary feedback $r_{t^*}$ for action $a_{t^*}$
                \State $\alpha_{t^*} \leftarrow \operatorname{sg}\!\left(|r_{t^*} - \pi_\theta(a_{t^*} \mid s_{t^*})|^{\gamma}\right)$
                \State Compute $\mathcal{L}(\tau) = \alpha_{t^*} \cdot \mathcal{L}_\mathrm{ent}(t^*)$ via Eq.~(\ref{eq:full_loss})
                \State Update $\theta \leftarrow \theta - \eta \nabla_\theta \mathcal{L}(\tau)$
            \EndIf
        \EndFor
        \Return Adapted policy $\pi_{\theta^\star}$
    \end{algorithmic}
\end{algorithm}

\section{Feedback-Efficient Test-Time Adaptation}
\begin{figure*}[t]
    \centering
    \includegraphics[width=\linewidth]{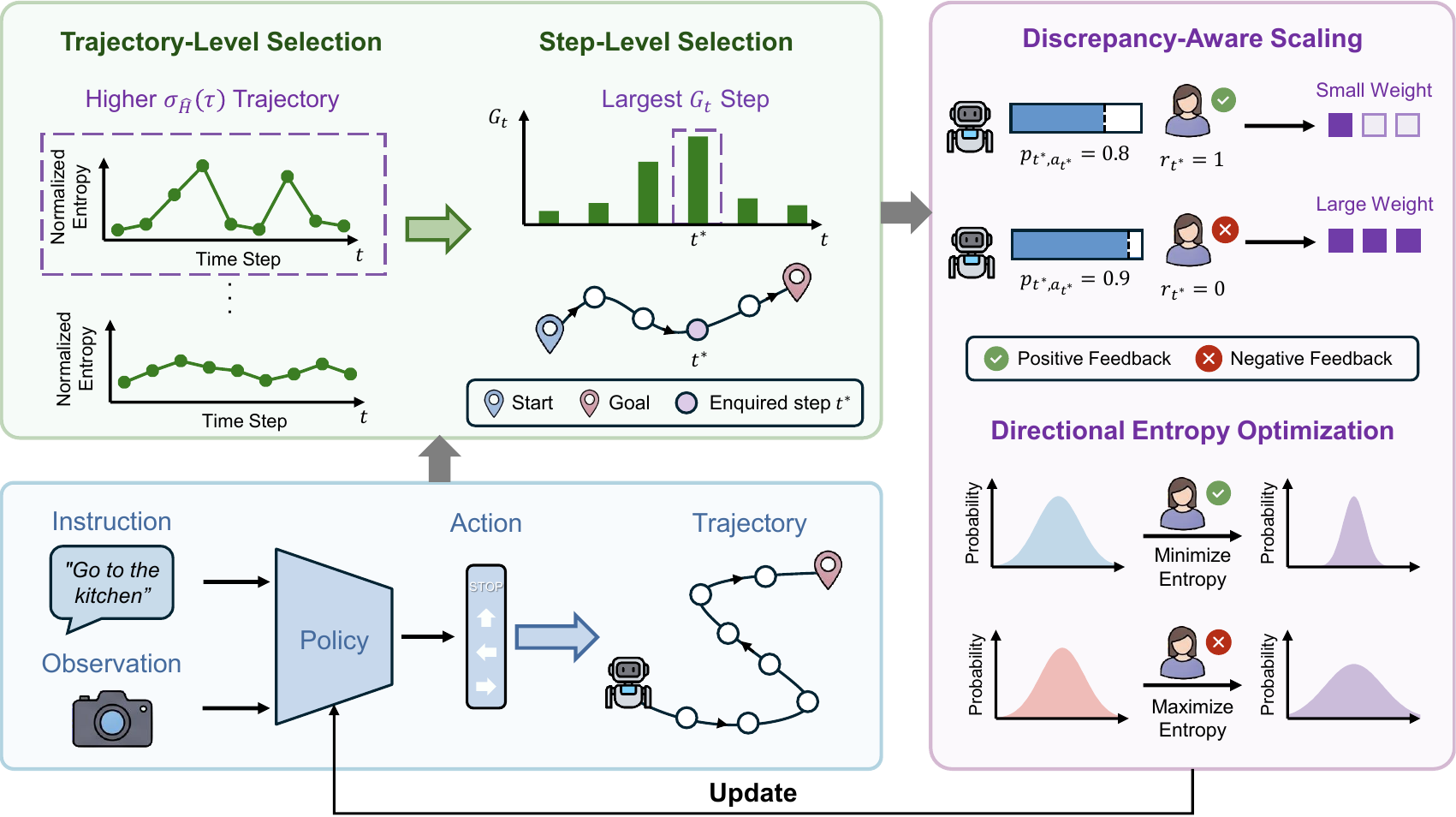}
    \caption{Overview of \methodname{}. For the trajectory executed by the agent, we hierarchically select informative samples by filtering trajectories with higher normalized-entropy standard deviation $\sigma_{\hat{H}}(\tau)$ and querying the step with the largest entropy-gradient saliency $G_{t}$. 
    The queried feedback is then used to update the policy via directional entropy optimization modulated by discrepancy-aware scaling, enabling efficient adaptation in novel environments.
    }
    \label{fig:method}
\end{figure*}

We propose \methodname{}, a \fullmethodname for VLN. At test time, after the agent executes a navigation trajectory in the target environment, \methodname{} first performs \textbf{Hierarchical Sample Selection} (\textit{c.f.} Section~\ref{sec:sample_selection}) to determine where human supervision should be used, including trajectory-level selection to identify informative trajectories and step-level selection to locate a promising decision point for feedback within each selected trajectory. \methodname{} then queries a binary feedback signal on the selected action (\textit{c.f.} Section~\ref{sec:prelimiary}). Based on the acquired feedback, \methodname{} finally performs \textbf{Feedback-Driven Policy Optimization} (\textit{c.f.} Section~\ref{sec:optimization}) to convert sparse supervision into discrepancy-aware step-level policy updates. In this way, \methodname{} concentrates limited human supervision on the informative decisions and enables efficient and reliable adaptation under distribution shift. Fig.~\ref{fig:method} provides an overview of our method, and Algorithm~\ref{alg:workflow} summarizes the complete procedure.

\subsection{Preliminary} 
\label{sec:prelimiary}
\noindent\textbf{Progress-Based Binary Feedback}. \methodname{} performs adaptation after each completed navigation episode, rather than updating the policy at every decision step. Accordingly, human feedback is queried only after the full trajectory is available. For each queried action, the human provides a binary signal indicating whether the action is beneficial for reaching the navigation goal. In the simulator, we instantiate this supervision with a progress-based oracle. Let $d_t$ denote the geodesic distance from the agent's position to the goal at time step $t$. For a non-stop action, the feedback is positive if the action reduces the agent-to-goal geodesic distance; for the stop action, the feedback is positive if the agent has successfully reached the goal. The binary reward is defined as:

\begin{equation}
r_t = 
\begin{cases}
\mathbbm{I}[d_t > d_{t+1}], & \text{if } a_t \neq \textsc{stop}, \\
\mathbbm{I}[d_T \leq d_{\text{th}}], & \text{if } a_t = \textsc{stop},
\end{cases}
\end{equation}
where $d_T$ is the final agent-to-goal geodesic distance and $d_{\text{th}}$ is the success threshold. $r_t = 1$ and $r_t = 0$ denote positive and negative feedback, respectively. This oracle enables reproducible evaluation, while local progress serves as an approximation to human judgment on action usefulness.

\begin{figure*}[t]
    \centering
    \begin{minipage}[t]{0.48\linewidth}
        \centering
        \includegraphics[width=\linewidth]{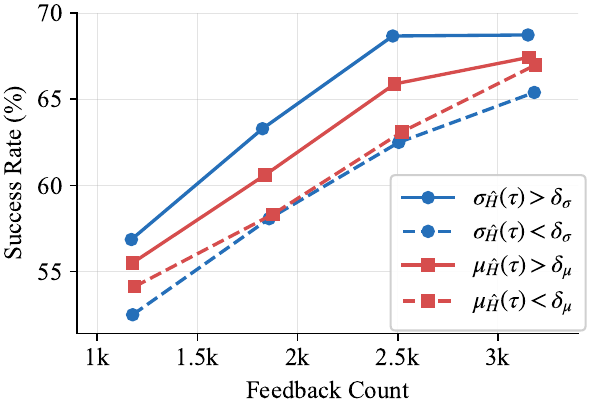}
        \caption{Trajectory-level selection criteria comparison. Solid/dashed lines select trajectories above/below threshold. $\delta_\sigma$ and $\delta_\mu$ are thresholds for $\sigma_{\hat{H}}(\tau)$ and $\mu_{\hat{H}}(\tau)$, respectively.}
        \label{fig:trajectory_selection_ablation}
    \end{minipage}
    \hfill
    \begin{minipage}[t]{0.48\linewidth}
        \centering
        \includegraphics[width=\linewidth]{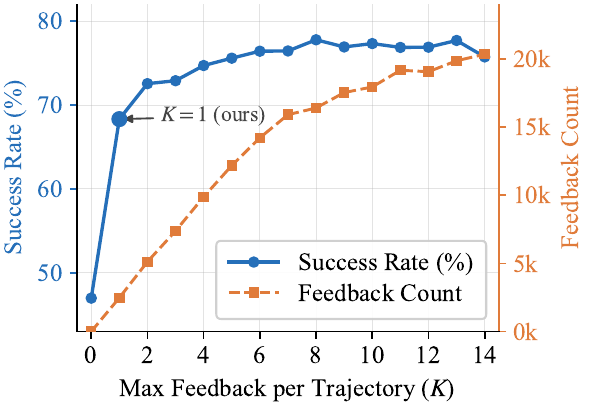}
        \caption{Effect of maximum number of feedback per trajectory ($K$) on success rate and total feedback count. Our method uses $K=1$ to balance feedback cost and performance.}
        \label{fig:action_num_ablation}
    \end{minipage}
\end{figure*}

\subsection{Hierarchical Sample Selection}
\label{sec:sample_selection}
Requesting human feedback on every action is impractical at test time. Under a limited feedback budget, we must answer two questions: \emph{which trajectories are worth querying}, and \emph{where to query within each selected trajectory}. We address these questions through a hierarchical sample selection strategy, whose design is motivated by two validation studies. As shown in Fig.~\ref{fig:trajectory_selection_ablation}, under comparable numbers of feedback, selecting trajectories with larger fluctuations in normalized policy entropy yields better adaptation gains, motivating our trajectory-level criterion. As shown in Fig.~\ref{fig:action_num_ablation}, although querying additional steps can further improve performance, the gains quickly diminish while the query cost continues to increase, indicating that one carefully selected step per trajectory provides the favorable trade-off. Guided by these observations, \methodname{} first performs \emph{Trajectory-Level Selection} to identify informative trajectories and then performs \emph{Step-Level Selection} to select a critical decision step for feedback querying.

\textbf{Trajectory-Level Selection.}
\label{sec:trajectory_selection}
Our goal is to identify the trajectories that are most likely to benefit from external feedback under a limited query budget. Since uncertainty is a natural indicator of adaptation potential, we measure the policy uncertainty along each trajectory using normalized entropy. Let $\pi_\theta(a_t \mid s_t)$ denote the softmax probability assigned to action $a_t$ at time step $t$, and let $|\mathcal{A}_t|$ denote the number of feasible actions at step $t$. The policy entropy and its normalized form are defined as:
\begin{equation}
\label{eq:norm_entropy}
H_t = -\sum_{a \in \mathcal{A}_t} \pi_\theta(a \mid s_t) \log \pi_\theta(a \mid s_t), \qquad
\hat{H}_t = \frac{H_t}{\log |\mathcal{A}_t|},
\end{equation}
where $\hat{H}_t\in[0,1]$ captures the relative uncertainty of the policy over the current action space. Based on the per-step normalized entropy within a trajectory $\tau$ of length $T$, we further define the mean and standard deviation of normalized entropy as:
\begin{equation}
\label{eq:traj_std}
\mu_{\hat{H}}(\tau) = \frac{1}{T}\sum_{t=1}^{T}\hat{H}_t,
\qquad
\sigma_{\hat{H}}(\tau) = \sqrt{\frac{1}{T}\sum_{t=1}^{T}\big(\hat{H}_t-\mu_{\hat{H}}(\tau)\big)^2 }.
\end{equation}
To compare the two criteria with closely matched feedback counts, we set their respective dynamic thresholds $\delta_\sigma$ and $\delta_\mu$ to the same $p$-th percentile of separate sliding windows, each containing the 150 most recent values of the corresponding criterion. As shown in Fig.~\ref{fig:trajectory_selection_ablation}, using the standard deviation $\sigma_{\hat{H}}(\tau)$ is more effective than using the mean entropy $\mu_{\hat{H}}(\tau)$. Intuitively, a large entropy standard deviation indicates that the agent alternates between confident and uncertain phases along the trajectory, suggesting the presence of localized unreliable decisions that can benefit more from sparse external correction. In contrast, high mean entropy indicates globally diffuse uncertainty, which is harder to correct with limited feedback. Based on this observation, we use $\sigma_{\hat{H}}(\tau)$ as the trajectory-level selection criterion:
\begin{equation}
S_\tau = \mathbbm{I}\bigl(\sigma_{\hat{H}}(\tau)>\delta_\sigma\bigr),
\end{equation}
where $\delta_\sigma$ is a dynamic threshold derived from recent trajectories (see Section~\ref{app:implementation_details}) and $\mathbbm{I}(\cdot)$ denotes the indicator function. Only trajectories with $S_\tau=1$ are forwarded to the next stage.

\textbf{Step-Level Selection.}
\label{sec:step_selection}
After selecting informative trajectories, the next question is where to query within each trajectory. Fig.~\ref{fig:action_num_ablation} shows that the first queried step yields the largest performance gain. While additional queried steps can further improve the success rate, their marginal benefits diminish quickly, whereas the feedback count increases nearly linearly. This makes a single carefully selected step the most cost-effective choice. Thus, \methodname{} queries only one step for each selected trajectory.

\blue{Given this one-query design, we further need to determine which step to query for correction. Intuitively, a useful query target should allow the feedback signal to guide policy optimization effectively. A natural alternative is to query the step with the highest entropy. However, high entropy only indicates that the policy is uncertain, and may simply correspond to a near-uniform distribution, where the entropy gradient with respect to the logits approaches zero. Instead, we select the step whose normalized entropy is most sensitive to the logits, using this sensitivity as a heuristic for identifying steps that may benefit from the correction.} We capture this property using the gradient norm of the normalized entropy with respect to the action logits, which we term \emph{entropy-gradient saliency}. Let $\mathbf{z}_t$ denote the action logits at step $t$. The saliency score is defined as:
\begin{equation}
\label{eq:saliency}
\begin{aligned}
G_t
&= \left\| \nabla_{z_t} \hat{H}_t \right\| \\
&= \frac{1}{\log|\mathcal{A}_t|} \sqrt{\sum_{a_j \in \mathcal{A}_t} \pi_\theta\left(a_j \mid s_t\right)^2 \left(\log \pi_\theta \left(a_j \mid s_t \right) + H_t \right)^2
}.
\end{aligned}
\end{equation}
For each selected trajectory, we rank all decision steps by $G_t$ and query feedback only for the executed action at the critical step $t^* = \arg\max_{t \in \tau}\, G_t$.

\subsection{Feedback-Driven Policy Optimization} \label{sec:optimization}
To effectively exploit sparse human feedback, we decompose each queried signal into two roles in policy adaptation. The prediction--feedback discrepancy modulates the update magnitude by reweighting how strongly a queried sample contributes to optimization, \blue{while the feedback label determines the direction of entropy optimization to encourage reinforcement or relaxation of the queried decision. In this way, feedback modulates the strength and determines the direction of entropy optimization.}

\textbf{Discrepancy-Aware Scaling.}
\label{sec:Adaptive_Gradient_Focusing}
Applying uniform-weight updates to all queried samples overlooks the fact that
different feedback samples are not equally informative for policy adaptation. To address this, we introduce a discrepancy-aware scaling factor according to the discrepancy between the current policy prediction and the queried feedback:
\begin{equation}
\label{eq:scaling_factor}
|r_t-\pi_\theta(a_t\mid s_t)|^\gamma,\qquad \gamma>0,
\end{equation}
where $\pi_\theta(a_t\mid s_t)$ denotes the probability assigned by the policy to the executed action $a_t$. When $\pi_\theta(a_t\mid s_t)$ is close to $r_t$, the current policy already agrees with the feedback, and the corresponding sample should contribute less to optimization. In contrast, when the discrepancy is large, the queried sample receives a larger relative weight, allowing the update to focus more on decision points where the current policy disagrees with the feedback signal.

\textbf{Directional Entropy Optimization.}
Different feedback signals imply different local adjustment needs. For  correct actions, the current decision should be reinforced; for incorrect ones, the current preference should be relaxed so that alternative actions can be explored. To realize this intuition, we use the policy entropy $H_t$ as the optimization objective and introduce a directional weight $w_t$:
\begin{equation}
\mathcal{L}_\mathrm{ent}(t) = w_t \cdot H_t, \qquad \mathrm{where}~~ w_t= \begin{cases} +1, & r_t=1,\\ -1, & r_t=0. \end{cases}
\end{equation}
When the feedback is positive, $w_t=+1$ encourages the local policy distribution to become more concentrated around the current action, thereby reinforcing the verified decision. When the feedback is negative, $w_t=-1$ encourages the current preference to be relaxed, allowing probability mass to shift toward alternative actions. 

\textbf{Full Optimization Objective.}
Combining the hierarchical sample filtering mechanism described in Section~\ref{sec:sample_selection}, the overall loss is defined as:
\begin{equation}
\label{eq:full_loss}
\mathcal{L}(\tau) =
\mathbbm{I}\!\big[\sigma_{\hat{H}}(\tau) > \delta_\sigma\big]
\cdot
\operatorname{sg}\!\left(|r_{t^*}-\pi_\theta(a_{t^*}\mid s_{t^*})|^\gamma\right)
\cdot
\mathcal{L}_\mathrm{ent}(t^*),
\end{equation}
where $t^* = \arg\max_{t \in \tau}\, G_t$ and $\operatorname{sg}(\cdot)$ denotes the stop-gradient operator, which treats the discrepancy term as a fixed weighting factor during optimization. This objective is computed only at the single queried step within trajectories that pass the trajectory-level filter. Here, the discrepancy term re-weights each queried sample according to its consistency with the received feedback, while $w_{t^*}$ determines whether the local decision is reinforced or relaxed. Together, these components enable \methodname{} to convert sparse binary feedback into effective local policy adaptation.

\begin{table*}[t]
\renewcommand{\tabcolsep}{6pt} 
\centering
\caption{Experimental results on REVERIE. The last column reports the average runtime per episode.}
\label{tab:reverie}
\small
\resizebox{\textwidth}{!}{
    \begin{tabular}{c *{11}{c}}
        \toprule
        \multirow{2}{*}{\textbf{Methods}}
        & \multicolumn{5}{c}{\textbf{Val Seen}}
        & \multicolumn{5}{c}{\textbf{Val Unseen}}
        & \multirow{2}{*}{Time(ms)} \\
        \cmidrule(lr){2-6} \cmidrule(lr){7-11}
        & OSR $\uparrow$ & SR $\uparrow$ & SPL $\uparrow$ & RGSPL $\uparrow$ & FC $\downarrow$
        & OSR $\uparrow$ & SR $\uparrow$ & SPL $\uparrow$ & RGSPL $\uparrow$ & FC $\downarrow$
        & \\
        \midrule
        \multicolumn{1}{l}{DUET~\cite{chen2022think}}
        & 73.86 & 71.75 & 63.94 & 51.14 & --
        & 51.07 & 46.98 & 33.73 & 23.03 & --
        & 156.5 \\
        \multicolumn{1}{l}{\hspace{0.5em}\textbullet~FSTTA~\cite{gao2024fastslow}}
        & 73.79 & 71.68 & 64.00 & 51.21 & -- 
        & 50.93 & 46.86 & 33.71 & 23.00 & -- 
        & 457.7 \\
        \multicolumn{1}{l}{\hspace{0.5em}\textbullet~FeedTTA~\cite{kim2025test}}
        & 82.83 & 77.28 & 65.39 & 54.28 & 1424 
        & 62.60 & 53.85 & 28.66 & 19.02 & 3522 
        & 433.1 \\
        \multicolumn{1}{l}{\hspace{0.5em}\textbullet~ATENA~\cite{ko2025active}}
        & 82.62 & 81.40 & 72.85 & 58.90 &  573
        & 65.78 & 61.95 & 42.73 & 28.39 &  2595
        & 321.3\\
        \rowcolor[HTML]{EEF4FF}
        \multicolumn{1}{l}{\hspace{0.5em}\textbullet~\textbf{\methodname{}}}
        & \textbf{84.31} & \textbf{82.29} & \textbf{73.75} & \textbf{59.01} & \textbf{519}
        & \textbf{73.27} & \textbf{68.33} & \textbf{50.55} & \textbf{33.15} & \textbf{2475} & 243.5 \\        
        
        \midrule
        \multicolumn{1}{l}{GOAT~\cite{Wang2024GOAT}}
        & 82.36 & 80.74 & 73.44 & 58.82 & --
        & 57.97 & 53.82 & 37.52 & 27.00 & --
        & 144.2 \\
        \multicolumn{1}{l}{\hspace{0.5em}\textbullet~FSTTA~\cite{gao2024fastslow}}
        & 82.85 & 80.67 & 72.65 & 58.11 & -- 
        & 57.54 & 53.48 & 36.76 & 25.80 & -- 
        & 461.1 \\
        \multicolumn{1}{l}{\hspace{0.5em}\textbullet~FeedTTA~\cite{kim2025test}}
        & 84.92 & 82.08 & 69.54 & 59.12 & 1424 
        & 65.12 & 56.75 & 33.68 & 26.11 & 3522 
        & 390.0 \\
        \multicolumn{1}{l}{\hspace{0.5em}\textbullet~ATENA~\cite{ko2025active}}
        & 84.68 & 83.02 & 76.46 & 61.46 & 803
        & 63.66 & 60.52 & 42.96 & 30.43 & 3012
        & 342.3 \\
        
        \rowcolor[HTML]{EEF4FF}
        \multicolumn{1}{l}{\hspace{0.5em}\textbullet~\textbf{\methodname{}}}
        & \textbf{90.02} & \textbf{87.65} & \textbf{77.98} & \textbf{62.14} & \textbf{533}
        & \textbf{72.75} & \textbf{68.65} & \textbf{54.22} & \textbf{38.18} & \textbf{2477} & 307.9\\
        
        \midrule
        \multicolumn{1}{l}{ScaleVLN~\cite{wang2023scaling}}
        & 73.30 & 70.98 & 60.86 & 44.30 & --
        & 61.43 & 54.25 & 41.28 & 25.00 & --
        & 187.7 \\
        \multicolumn{1}{l}{\hspace{0.5em}\textbullet~FSTTA~\cite{gao2024fastslow}}
        & 73.23 & 70.91 & 60.82 & 44.25 & -- 
        & 61.45 & 54.15 & 41.24 & 25.00 & -- 
        & 544.7 \\
        \multicolumn{1}{l}{\hspace{0.5em}\textbullet~FeedTTA~\cite{kim2025test}}
        & 78.71 & 72.10 & 55.53 & 45.92 & 1424 
        & 71.71 & 56.18 & 34.85 & 23.83 & 3522 
        & 437.5 \\
        \multicolumn{1}{l}{\hspace{0.5em}\textbullet~ATENA~\cite{ko2025active}}
        & 77.93 & 76.04 & 66.01 & 48.06 & 965
        & 68.49 & 63.32 & 49.08 & 29.20 & 2870
        & 344.4 \\
        
        \rowcolor[HTML]{EEF4FF}
        \multicolumn{1}{l}{\hspace{0.5em}\textbullet~\textbf{\methodname{}}}
        & \textbf{83.02} & \textbf{80.86} & \textbf{70.76} & \textbf{51.67} & \textbf{562} 
        & \textbf{75.50} & \textbf{69.84} & \textbf{56.61} & \textbf{33.52} & \textbf{2465} & 300.0\\
        
        \bottomrule
    \end{tabular}
}
\end{table*}

\begin{table}[t]
    \centering
    \renewcommand{\tabcolsep}{1pt}
    \caption{Experimental results on GSA-R2R.}
    \label{tab:gsa-r2r}
    \resizebox{\linewidth}{!}{
        \begin{tabular}{c *{8}{c}}
            \toprule
            \multirow{2}{*}{\textbf{Methods}}
            & \multicolumn{4}{c}{\textbf{Test Residential Basic}}
            & \multicolumn{4}{c}{\textbf{Test Non-residential Scene}} \\
            \cmidrule(lr){2-5} \cmidrule(lr){6-9}
            & SR $\uparrow$ & SPL $\uparrow$ & NDTW $\uparrow$ & NE $\downarrow$
            & SR $\uparrow$ & SPL $\uparrow$ & NDTW $\uparrow$ & NE $\downarrow$ \\
            \midrule
            \multicolumn{1}{l}{GR-DUET~\cite{hong2025general}}
            & 69 & 64 & 71 & 3.09
            & 47 & 42 & 54 & 5.55  \\
            \multicolumn{1}{l}{\hspace{0.5em}\textbullet~FSTTA~\cite{gao2024fastslow}}
            & 69 & 62 & 70 & 3.14
            & 48 & 42 & 53 & 5.46  \\
            \multicolumn{1}{l}{\hspace{0.5em}\textbullet~FeedTTA~\cite{kim2025test}}
            & 69 & 64 & 72 & 3.07
            & 48 & 43 & 55 & 5.43  \\
            \multicolumn{1}{l}{\hspace{0.5em}\textbullet~ATENA~\cite{ko2025active}}
            & 67 & 64 & 71 & 3.34 
            & 48 & 40 & 50 & 5.45  \\
            
            \rowcolor[HTML]{EEF4FF}   \multicolumn{1}{l}{\hspace{0.5em}\textbullet~\textbf{\methodname{}}}
            & \textbf{71} & \textbf{67} & \textbf{74} & \textbf{2.93}
            & \textbf{50} & \textbf{44} & \textbf{55} & \textbf{4.94} \\
            
            \midrule
            \multicolumn{1}{l}{BEVBert~\cite{an2023bevbert}}
            & 58 & 45 & 54 & 4.11 
            & 39 & 27 & 36 & 6.69  \\
            \multicolumn{1}{l}{\hspace{0.5em}\textbullet~FSTTA~\cite{gao2024fastslow}}
            & 58 & 44 & 53 & 4.22
            & 40 & 28 & 37 & 6.50  \\
            \multicolumn{1}{l}{\hspace{0.5em}\textbullet~FeedTTA~\cite{kim2025test}}
            & 59 & 44 & 53 & 4.09
            & 39 & 26 & 35 & 6.70  \\
            \multicolumn{1}{l}{\hspace{0.5em}\textbullet~ATENA~\cite{ko2025active}}
            & 58 & 44 & 53 & 4.20 
            & 40 & 28 & 37 & 6.49  \\
            
            \rowcolor[HTML]{EEF4FF}  \multicolumn{1}{l}{\hspace{0.5em}\textbullet~\textbf{\methodname{}}}
            & \textbf{59} & \textbf{45} & \textbf{54} & \textbf{3.91}
            &  \textbf{42} & \textbf{28} & \textbf{37} & \textbf{5.75}  \\
            \bottomrule
        \end{tabular}
    }
\end{table}

\section{Experiments}
\subsection{Benchmark and Metrics}
\label{sec:metrics}
We evaluate our method on four benchmarks: REVERIE~\cite{qi2020reverie}, R2R~\cite{anderson2018vision}, GSA-R2R~\cite{hong2025general}, and R2R-CE~\cite{krantz_vlnce_2020}. REVERIE requires the agent to navigate to a target region and identify the referred object based on a high-level instruction. R2R focuses on fine-grained instruction following in discrete indoor environments. GSA-R2R extends R2R with more diverse environments and instruction styles to evaluate agent adaptability in both in-distribution and out-of-distribution settings. R2R-CE transfers the R2R task to continuous environments, requiring low-level action execution. We report \textbf{SR} (Success Rate), \textbf{OSR} (Oracle Success Rate), \textbf{SPL} (Success rate weighted by Path Length), \textbf{NDTW} (Normalized Dynamic Time Warping), \textbf{TL} (trajectory length) and \textbf{NE} (Navigation Error). For REVERIE, we additionally report \textbf{RGS} and \textbf{RGSPL}, which measure remote object grounding success without and with path efficiency weighting. We also introduce \textbf{FC} (Feedback Count) to quantify the total number of human feedback queries during evaluation, directly reflecting the annotation cost of test-time adaptation process. R2R results are provided in the Appendix~\ref{app:r2r_result}.

\begin{table}[t]
    \centering
    \renewcommand{\tabcolsep}{2.0pt}
    \caption{Experimental results on R2R-CE.}
    \label{tab:r2rce}
    \resizebox{\linewidth}{!}{
        \begin{tabular}{c *{8}{c}}
            \toprule
            \multirow{2}{*}{\textbf{Methods}}
            & \multicolumn{4}{c}{\textbf{Val Seen}}
            & \multicolumn{4}{c}{\textbf{Val Unseen}} \\
            \cmidrule(lr){2-5} \cmidrule(lr){6-9}
            & OSR $\uparrow$ & SR $\uparrow$ & SPL $\uparrow$ & NE $\downarrow$
            & OSR $\uparrow$ & SR $\uparrow$ & SPL $\uparrow$ & NE $\downarrow$ \\
            \midrule
            \multicolumn{1}{l}{ETPNav~\cite{an2024etpnav}}
            & 71 & 64 & 57 & 4.12
            & 64 & 57 & 49 & 4.72 \\
            \multicolumn{1}{l}{\hspace{0.5em}\textbullet~FSTTA~\cite{gao2024fastslow}}
            & 70 & 63 & 56 & 4.19
            & 63 & 55 & 47 & 4.86  \\
            \multicolumn{1}{l}{\hspace{0.5em}\textbullet~FeedTTA~\cite{kim2025test}}
            & 72 & 65 & 58 & 4.06
            & 64 & 57 & 49 & 4.73  \\
            \multicolumn{1}{l}{\hspace{0.5em}\textbullet~ATENA~\cite{ko2025active}}
            & 70  & 65  & 59  & 4.06 
            & 63  & 57  & 50  & 4.69  \\
            \rowcolor[HTML]{EEF4FF}
            \multicolumn{1}{l}{\hspace{0.5em}\textbullet~\textbf{\methodname{}}}
            & \textbf{72} & \textbf{66} & \textbf{59} & \textbf{4.00}
            & \textbf{65} & \textbf{59} & \textbf{51} & \textbf{4.52} \\
            \midrule
            \multicolumn{1}{l}{BEVBert~\cite{an2023bevbert}}
            & 78  & 71  & 63  & 3.22 
            & 66  & 58  & 49  & 4.53  \\
            \multicolumn{1}{l}{\hspace{0.5em}\textbullet~FSTTA~\cite{gao2024fastslow}}
            & 75 & 68 & 60 & 3.52
            & 67 & 59 & 49 & 4.59  \\
            \multicolumn{1}{l}{\hspace{0.5em}\textbullet~FeedTTA~\cite{kim2025test}}
            & 77 & 72 & 64 & 3.21
            & 67 & 58 & 49 & 4.47  \\
            \multicolumn{1}{l}{\hspace{0.5em}\textbullet~ATENA~\cite{ko2025active}}
            & 77  & 71  & 64  & 3.24 
            & 67  & 60  & 48  & 4.34  \\
            \rowcolor[HTML]{EEF4FF}
            \multicolumn{1}{l}{\hspace{0.5em}\textbullet~\textbf{\methodname{}}}
            & \textbf{79}  & \textbf{73}  & \textbf{64}  & \textbf{3.08}
            & \textbf{69}  & \textbf{61}  & \textbf{50}  & \textbf{4.20}  \\
            \bottomrule
        \end{tabular}
    }
\end{table}

\subsection{Main Navigation Results}
\textbf{Results on REVERIE.}
Table~\ref{tab:reverie} reports the comparative results on the REVERIE~\cite{qi2020reverie} benchmark. We conduct adaptation using DUET~\cite{chen2022think}, GOAT~\cite{Wang2024GOAT}, and ScaleVLN~\cite{wang2023scaling} as base models. Across all base models and data splits, \methodname{} consistently outperforms ATENA~\cite{ko2025active}, the current state-of-the-art test-time adaptation method for navigation. In particular, on the more challenging val unseen split, \methodname{} achieves relative SPL improvements over ATENA of 18.30\%, 26.21\%, and 15.34\% for DUET, GOAT, and ScaleVLN, respectively. Meanwhile, \methodname{} also achieves lower feedback count (FC) across all settings, \blue{as the hierarchical selection strategy filters  informative trajectories and queries critical navigation step within each selected trajectory.} Furthermore, \methodname{} requires consistently less execution time per episode than ATENA, demonstrating that the proposed method improves both effectiveness and efficiency.

\textbf{Results on GSA-R2R.}
We further evaluate \methodname{} on the scene-diverse GSA-R2R~\cite{hong2025general} benchmark. As shown in Table~\ref{tab:gsa-r2r}, ATENA brings only limited gains and even degrades the performance of GR-DUET on the residential split. In contrast, \methodname{} delivers consistent improvements across all settings and achieves up to a 10.00\% relative improvement in SPL over ATENA on the more challenging non-residential split. This suggests that \methodname{} remains effective even when deployment scenarios differ substantially from the training distribution.

\textbf{Results on R2R-CE.}
Unlike the discrete navigation settings above, R2R-CE~\cite{krantz_vlnce_2020} requires the agent to execute low-level continuous control commands. As shown in Table~\ref{tab:r2rce}, \methodname{} achieves the best results with both ETPNav~\cite{an2024etpnav} and BEVBert~\cite{an2023bevbert}. On the val unseen split, it achieves relative SR improvements over ATENA of 3.51\% and 1.67\% with ETPNav and BEVBert, respectively, demonstrating consistent gains in continuous environments.


\begin{table}[t!]
    \setlength{\tabcolsep}{4.0pt} 
    \centering
    \caption{Ablation study of Trajectory-Level Selection (TLS), Step-Level Selection (SLS), and Discrepancy-Aware Scaling (DAS) on the REVERIE Val Unseen split. Baseline denotes the unadapted model.}
    \label{tab:ablation}
    \resizebox{\linewidth}{!}{
        \begin{tabular}{*{8}{c}}
            \toprule
            \multicolumn{3}{c}{\textbf{Module}} & \multicolumn{5}{c}{\textbf{Val Unseen}} \\ 
            \cmidrule(lr){1-3} \cmidrule(lr){4-8}
            TLS & SLS & DAS & SR $\uparrow$ 
            & OSR $\uparrow$ & SPL $\uparrow$
            & RGSPL $\uparrow$ & RGS $\uparrow$  \\
            \midrule
            \multicolumn{3}{c}{Baseline} & 46.98 & 51.07 & 33.73 & 23.03 & 32.15 \\
            \midrule
            \checkmark & - & - & 55.54 & 64.21 & 36.10 & 24.12 & 37.06 \\ 
            - & \checkmark & - & 61.41 & 69.01 & 43.32 & 29.14 & 41.47  \\
            - & - & \checkmark & 56.83 & 62.52 & 37.63 & 25.11 & 37.91  \\
            - & \checkmark & \checkmark & 63.19 & 67.52 & 46.86 & 31.32 & 42.34 \\
            \checkmark & - & \checkmark & 57.66 & 62.69 & 41.20 & 27.77 & 38.90 \\
            \checkmark & \checkmark & - & 65.46 & 72.05 & 48.42 & 32.62 & 44.08 \\
            \checkmark & \checkmark & \checkmark & 
            \textbf{68.33} & \textbf{73.27} & \textbf{50.55} & \textbf{33.15} & \textbf{45.08} \\
            \bottomrule
        \end{tabular}
    }
\end{table}

\subsection{Ablation Study}
To verify the effectiveness of each component in our method, we conduct ablation studies on the REVERIE Val Unseen split, with results reported in Table~\ref{tab:ablation}. To ensure a fair comparison under an equal feedback budget, the ablations of Trajectory-Level Selection (TLS) and Step-Level Selection (SLS) replace the respective selection method with uniform random sampling while keeping the total number of feedback queries identical. \blue{Disabling Discrepancy-Aware Scaling (DAS) assigns unit weight to each queried sample. Relative to the full method, replacing TLS with random trajectory sampling leads to a performance drop across all reported metrics, supporting the effectiveness of trajectory selection based on the standard deviation of normalized entropy. Replacing SLS with random step sampling also decreases performance, indicating that entropy-gradient saliency helps make effective use of sparse feedback. Removing DAS while retaining TLS and SLS also decreases performance, supporting the benefit of weighting each queried sample according to the prediction--feedback discrepancy. The variants with only TLS, SLS, or DAS enabled also outperform the unadapted baseline, showing that adaptation remains effective when only one of the three components is enabled.} These results collectively demonstrate that our method efficiently exploits sparse human feedback to enable effective agent adaptation in unseen environments.


                


\begin{table}[t!]
    \setlength{\tabcolsep}{3.0pt} 
    \centering
    \caption{Experimental results of cross-dataset adaptation. Models are trained on one dataset and adapted on the other.}
    \label{tab:cross_dataset}
    \resizebox{\linewidth}{!}{
        \begin{tabular}{lccccc}
        \toprule
        \multirow{2}{*}{\makecell[c]{\textbf{Data (Train $\rightarrow$ Test)}}} 
        & \multirow{2}{*}{\makecell[l]{\textbf{Methods}}} 
        & \multicolumn{4}{c}{\textbf{Val Unseen}} \\
        \cmidrule(lr){3-6}
        & & SR~$\uparrow$ & SPL~$\uparrow$ & NE~$\downarrow$ & TL~$\downarrow$ \\
        \midrule

        \multirow{3}{*}{R2R $\rightarrow$ REVERIE} 
        & DUET~\cite{chen2022think} 
        & 24.91 & 20.00 & 7.22 & \textbf{14.67} \\
                
        & \multicolumn{1}{l}{\hspace{0.5em}\textbullet~ATENA} 
        & 29.42 & 19.43 & 6.75 & 24.45 \\
        
        & \multicolumn{1}{l}{\hspace{0.5em}\textbullet~\methodname{}} 
        & \textbf{40.02} & \textbf{23.53} & \textbf{5.20} & 39.09 \\
        \midrule

        \multirow{3}{*}{REVERIE $\rightarrow$ R2R} 
        & DUET~\cite{chen2022think} 
        & 29.84 & 9.03 & 7.28 & 44.20 \\
        
        & \multicolumn{1}{l}{\hspace{0.5em}\textbullet~ATENA} 
        & 31.25 & 8.41 & 6.98 & 48.40 \\
        
        & \multicolumn{1}{l}{\hspace{0.5em}\textbullet~\methodname{}} 
        & \textbf{36.91} & \textbf{19.06} & \textbf{6.23} & \textbf{34.65} \\
        \bottomrule
        \end{tabular}
    }
\end{table}

\begin{table}[t]
    \centering
    \caption{Experimental results in real-world environments.}
    \label{tab:real_world_navigation}
    \renewcommand{\arraystretch}{1.10}
    \setlength{\tabcolsep}{12pt}
    \resizebox{\linewidth}{!}{
        \begin{tabular}{@{}lcccc@{}}
            \toprule
            \multirow{2}{*}{Method}
            & \multicolumn{2}{c}{Office}
            & \multicolumn{2}{c}{Laboratory} \\
            \cmidrule(lr){2-3} \cmidrule(lr){4-5}
            & SR $\uparrow$ & NE $\downarrow$
            & SR $\uparrow$ & NE $\downarrow$ \\
            \midrule
            BEVBert~\cite{an2023bevbert} & 40 & 4.30 & 30 & 4.20 \\
            \hspace{0.5em}\textbullet~\textbf{\methodname{}}
            & \textbf{55} & \textbf{3.87}
            & \textbf{40} & \textbf{4.03} \\
            \bottomrule
        \end{tabular}
    }
\end{table}

\subsection{More Discussions}
\textbf{Comparison of Step-Level Selection Methods.}
Table~\ref{tab:act_selection_method} compares entropy-gradient saliency against querying feedback at the step with highest normalized entropy. Under an equal budget, entropy gradient saliency outperforms the entropy-based selection method across all metrics, yielding an 8.97\% relative improvement in SPL on the REVERIE Val Unseen split. A possible reason is that time steps with high entropy may correspond to policy outputs that approach a uniform distribution. Negative feedback at such steps provides a weak entropy-maximization signal, as the entropy gradient with respect to the logits approaches zero near a uniform distribution. By contrast, selecting the step with the largest entropy-gradient saliency $G_t$ directs feedback toward steps where normalized entropy is more sensitive to logit changes, which may facilitate feedback-driven policy adjustment.

\begin{figure}[t]
    \centering
    \includegraphics[width=0.92\linewidth]{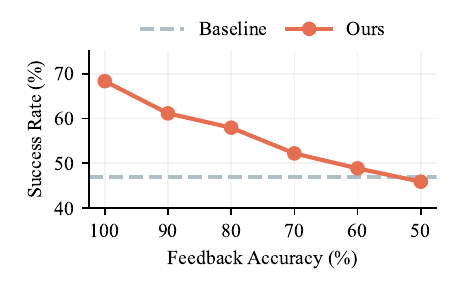}
    \caption{Effect of noisy binary feedback on navigation performance in the REVERIE Val Unseen split.}
    \label{fig:noise_sr}
\end{figure}

\textbf{Robustness of the Method.}
\label{sec:robustness}
In practical applications, binary feedback provided by humans is rarely perfectly accurate. To evaluate the robustness of our method under noisy feedback, we simulate noise by randomly flipping feedback labels with varying probabilities, gradually reducing the feedback accuracy from 100\% to 50\%. As shown in Fig.~\ref{fig:noise_sr}, the navigation success rate decreases as the noise level increases. Even at 60\% feedback accuracy, our method still outperforms the unadapted baseline. Among the tested noise levels, performance falls below the baseline only when feedback becomes random (50\% accuracy). These results demonstrate that \methodname{} can tolerate substantial simulated label noise while retaining an advantage over the unadapted baseline.

\begin{table*}[t]
\setlength{\tabcolsep}{8.0pt} 
\centering
\caption{Comparison of step-level selection methods on the REVERIE dataset.}
\label{tab:act_selection_method}
    \resizebox{0.9\textwidth}{!}{
    \begin{tabular}{c *{4}{c} *{4}{c}}
        \toprule
        \multirow{2}{*}{\textbf{Selection Method}}
        & \multicolumn{4}{c}{\textbf{Val Seen}}
        & \multicolumn{4}{c}{\textbf{Val Unseen}}\\
        \cmidrule(lr){2-5} \cmidrule(lr){6-9}
        & OSR$\uparrow$ & SR$\uparrow$ & SPL$\uparrow$ & RGSPL$\uparrow$
        & OSR$\uparrow$ & SR$\uparrow$ & SPL$\uparrow$ & RGSPL$\uparrow$ \\
        \midrule
        Entropy
        & 81.19 & 79.03 & 71.31 & 57.41
        & 69.63 & 63.97 & 46.39 & 31.13 \\
        Entropy-Gradient Saliency
        & \textbf{84.31} & \textbf{82.29} & \textbf{73.75} & \textbf{59.01} 
        & \textbf{73.27} & \textbf{68.33} & \textbf{50.55} & \textbf{33.15} \\
        \bottomrule
    \end{tabular}
    }
\end{table*}

\begin{figure}[t]
    \centering
    \includegraphics[width=0.92\linewidth]{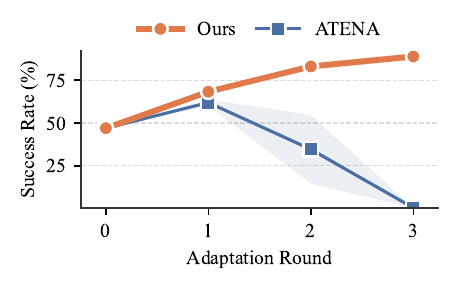}
    \caption{Continual adaptation results on REVERIE Unseen split, averaged over 3 random seeds.}
    \label{fig:continual_adapt}
\end{figure}

\textbf{Continual Adaptation.}
We evaluate continual adaptation by progressively adapting the model over multiple rounds and measuring performance after each round, starting from the unadapted model (Round 0). As shown in Fig.~\ref{fig:continual_adapt}, ATENA improves over the baseline after the first round but degrades in subsequent rounds, eventually falling below the unadapted performance. \blue{In contrast, \methodname{} improves in each of the three adaptation rounds. This contrast may be related to the granularity of supervision: ATENA uses the same trajectory-level feedback signal across all steps, potentially reinforcing suboptimal actions or suppressing locally correct ones. By contrast, \methodname{} computes the adaptation loss at a selected step using feedback on the corresponding action, aligning the supervision granularity with the decision being optimized.}

\textbf{Cross-Dataset Adaptation.}
We evaluate cross-dataset adaptation by training the model on one dataset and then adapting it to the other. Table~\ref{tab:cross_dataset} reports cross-dataset generalization under two transfer directions: R2R$\rightarrow$REVERIE and REVERIE$\rightarrow$R2R. Compared with the competing methods, \methodname{} shows strong transferability across datasets and tasks, achieving better adaptation performance. These results confirm that our adaptation strategy is not dataset-specific but generalizes robustly across heterogeneous task distributions.

\blue{
\textbf{Real-Robot Navigation Results.}
We evaluate \methodname{} on a physical robot in an office and a laboratory, with 20 navigation episodes per environment. We use BEVBert~\cite{an2023bevbert} as the baseline and retain the same settings as in simulation. As shown in Table~\ref{tab:real_world_navigation}, \methodname{} improves success rate by 15 and 10 percentage points in the office and laboratory, respectively, while also reducing navigation error in both environments. These results provide initial evidence that \methodname{} can improve navigation performance in physical environments beyond simulation.
}

\section{Human--Simulator Feedback Agreement}
\blue{To assess how closely the progress-based simulator feedback reflects human judgments of action usefulness, we conduct a human annotation study on 90 feedback queries from the REVERIE Val Unseen split. Three annotators each label a disjoint set of 30 queries, so each query receives one human annotation. For each query, the annotator is shown the navigation instruction, the complete observation sequence along the trajectory, the observations immediately before and after the queried action, and the action itself. The annotator then assigns a binary label indicating whether the action is helpful for completing the task. Human annotations agree with the simulator feedback on 78.9\% of queries, with agreement rates of 81.8\% for simulator-positive queries and 76.1\% for simulator-negative queries. The label distributions are approximately balanced: 48.9\% of simulator labels are positive, while annotators judge 52.2\% of actions as helpful. These results provide empirical support for using the simulator feedback as a proxy for human judgments in this setting. The annotation interface is shown in Fig.~\ref{fig:label_page} in the appendix.}

\section{Implementation Details}
\label{app:implementation_details}
To simulate real-world deployment conditions, we adopt an online test-time adaptation paradigm with a batch size of 1. All adaptation methods are evaluated three times with distinct random orderings of test samples, and we report the mean performance across runs. We evaluate our method against several representative VLN architectures, including DUET~\cite{chen2022think}, GOAT~\cite{Wang2024GOAT}, ScaleVLN~\cite{wang2023scaling}, ETPNav~\cite{an2024etpnav}, and BEVBert~\cite{an2023bevbert}. All baseline results are reproduced using the official implementations. During adaptation, we perform full-parameter updates with learning rates chosen from $\{6\times10^{-6}, 2\times10^{-6}, 6\times10^{-7}\}$. A gradient update is triggered at the end of an episode only if the standard deviation of the normalized entropy $\sigma_{\hat{H}}(\tau)$ across the trajectory exceeds a dynamic threshold $\delta_\sigma$. Specifically, we maintain a sliding window of the $W=150$ most recent standard deviations. Before the window is filled, all episodes pass the filter since feedback is valuable when the agent first encounters a novel environment. On REVERIE, R2R and R2R-CE, $\delta_\sigma$ is set to the 70th percentile of this window for Val Seen and the 30th percentile for Val Unseen. On GSA-R2R, $\delta_\sigma$ is set to the 30th percentile for all evaluations. These settings balance feedback cost and navigation performance. For episodes that pass this filter, the update is applied at the navigation step exhibiting the highest entropy-gradient saliency $G_t$. The discrepancy-aware scaling factor $\gamma$ in Eq.~(\ref{eq:scaling_factor}) is set to 0.6. The results of FSTTA~\cite{gao2024fastslow}, FeedTTA~\cite{kim2025test}, and ATENA~\cite{ko2025active} are reproduced using its officially released codebase. All experiments are conducted on a single NVIDIA Tesla A800 GPU.

\begin{figure*}[t]
    \centering
    \begin{minipage}[t]{0.48\linewidth}
        \centering
        \includegraphics[width=\linewidth]{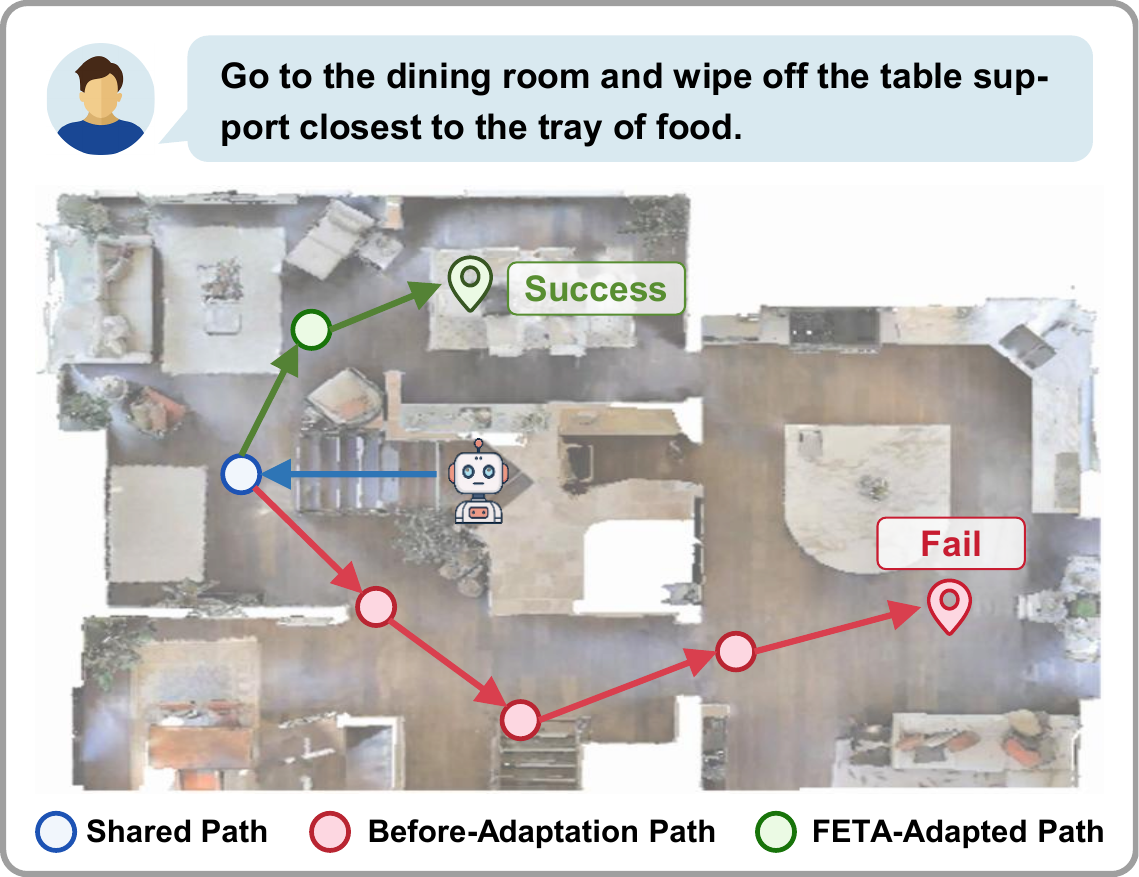}
        \label{fig:qualitative_analysis_1}
    \end{minipage}
    \hfill
    \begin{minipage}[t]{0.48\linewidth}
        \centering
        \includegraphics[width=\linewidth]{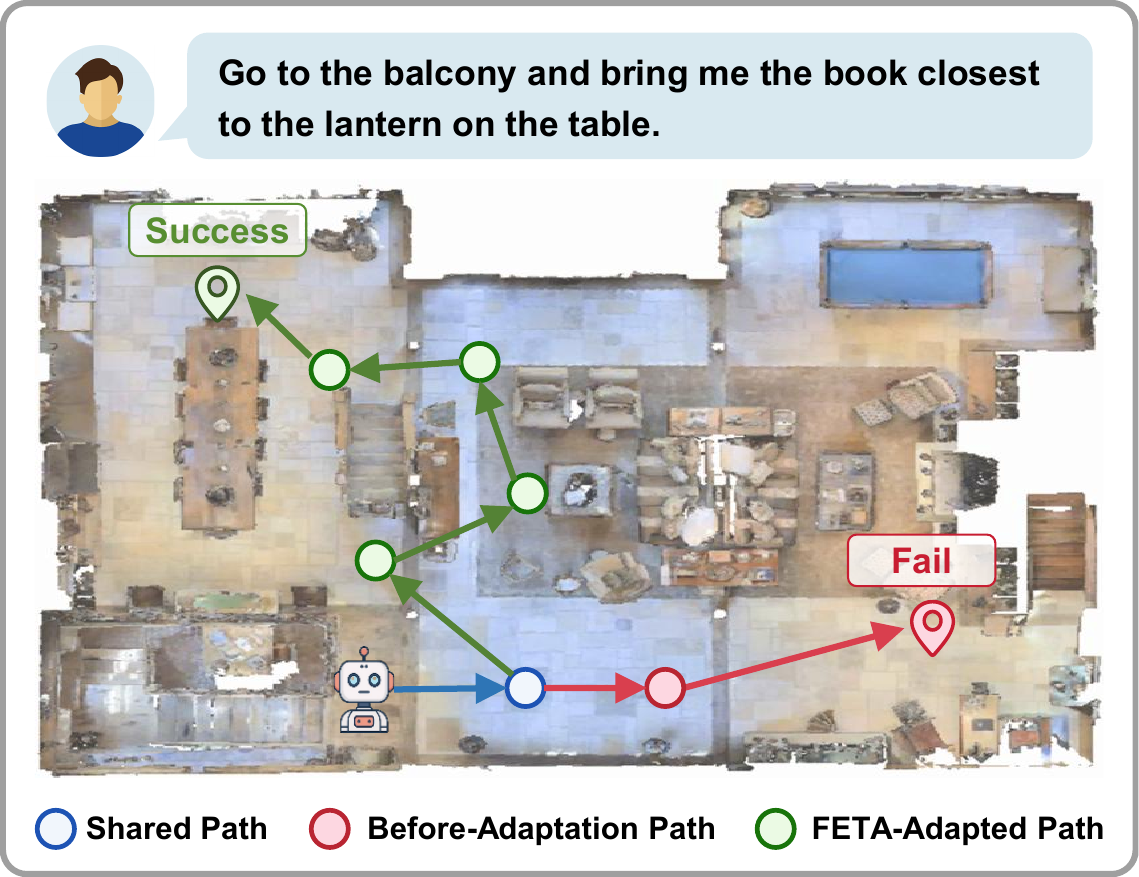}
        \label{fig:qualitative_analysis_2}
    \end{minipage}
    \caption{Qualitative comparison of navigation trajectories between the base model and our test-time adaptation method (\methodname{}). Red waypoints indicate the trajectory predicted by the base model, while green waypoints denote the trajectory predicted after applying \methodname{} at test time. Blue waypoints are the shared prefix visited by both methods before their trajectories diverge.}
    \label{fig:qualitative_analysis}
\end{figure*}

\begin{figure*}[t]
    \centering
    \includegraphics[width=\textwidth]{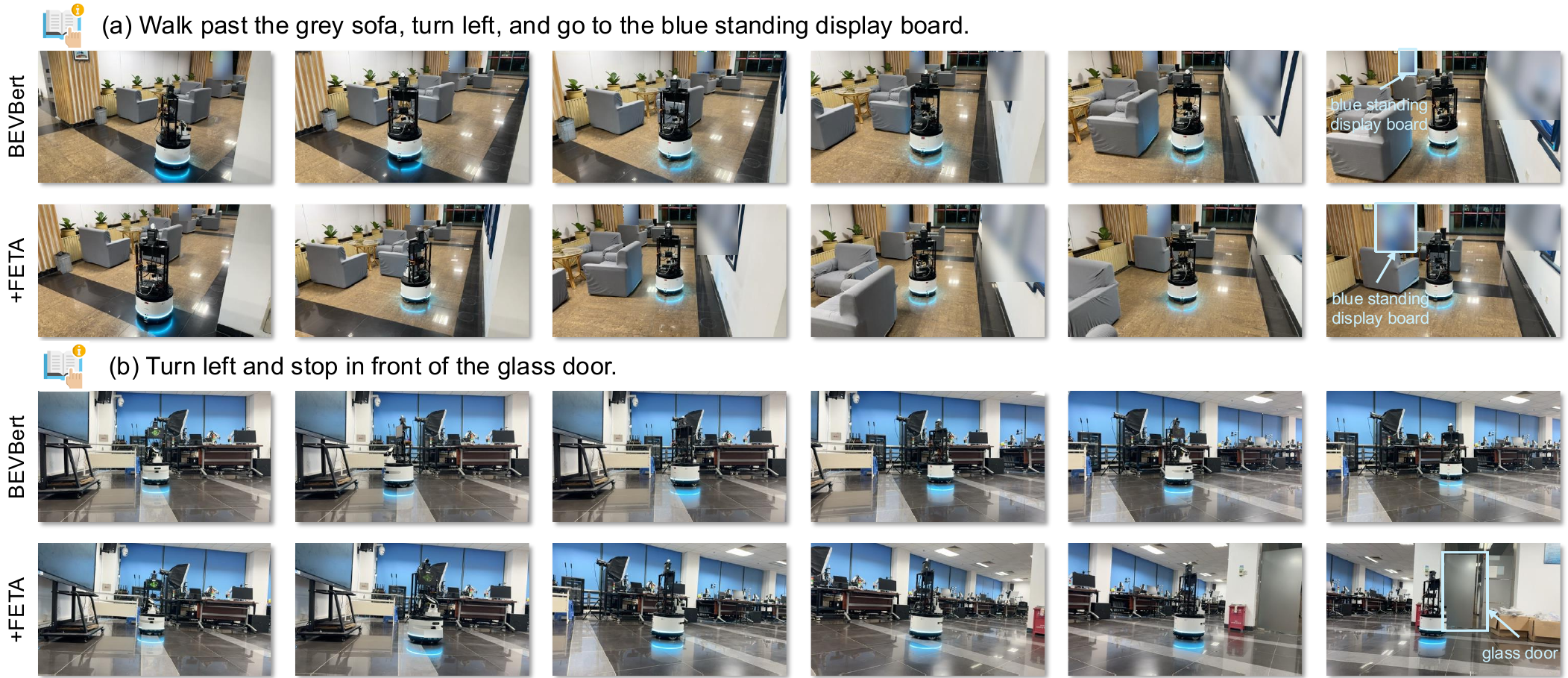}
    \caption{Real-robot navigation examples. (a) BEVBert fails to reach the blue display board, while \methodname{} proceeds past the sofas and stops near the board. (b) BEVBert remains near the starting location, while \methodname{} successfully stops in front of the glass door.}
    \label{fig:real_traj}
    \vspace{-3pt}
\end{figure*}

\section{Qualitative Analysis}
Fig.~\ref{fig:qualitative_analysis} visualizes the navigation trajectories of the base policy (DUET~\cite{chen2022think}) and the policy adapted by \methodname{} on two representative episodes from the REVERIE validation unseen split. In the left panel of Fig.~\ref{fig:qualitative_analysis}, the instruction asks the agent to go to the dining room and identify the table support closest to the tray of food. The unadapted model navigates toward a plausible dining area containing a round table, but ultimately stops at an incorrect location. After adaptation, \methodname{} steers the agent toward the correct dining room, yielding a trajectory that better matches the instructed destination. In the right panel of Fig.~\ref{fig:qualitative_analysis}, the instruction requires the agent to navigate to the balcony and locate the book closest to the lantern on the table. The unadapted model terminates in the living room rather than reaching the requested balcony. After adaptation, \methodname{} leads the agent to the balcony specified by the instruction. \blue{Across both examples, a consistent pattern emerges: the unadapted policy makes a plausible but erroneous decision at a branching point and subsequently fails to reach the target, whereas the adapted policy follows a route toward the instructed destination. 
}

\blue{The improvement in reaching the instructed destination is also reflected in the real-robot examples in Fig.~\ref{fig:real_traj}, where BEVBert serves as the base policy. In example (a), the adapted policy proceeds past the sofas and stops near the blue display board, which the baseline fails to reach. In example (b), the adapted policy reaches the specified glass door, while the baseline remains near the starting location. Together, the simulation and real-robot examples provide qualitative evidence that \methodname{} can help navigation policies follow routes toward instruction-specified destinations in both simulated and physical environments.}    

\section{Conclusion}
We propose \methodname{}, a \fullmethodname{} for Vision-and-Language Navigation. To address the inefficient feedback utilization and mismatched optimization granularity of prior methods, \methodname{} performs hierarchical sample selection that filters informative trajectories via the standard deviation of normalized entropy and \blue{selects a critical step for targeted feedback} via entropy-gradient saliency. The queried feedback is then converted into a fine-grained step-level update through a discrepancy-aware entropy optimization objective. Extensive experiments show that \methodname{} achieves state-of-the-art performance with fewer human queries and lower per-episode latency.




\bibliographystyle{IEEEtran}
\bibliography{ref}

\begin{IEEEbiography}[{\includegraphics[width=1.0 in,height=1.35in, clip,keepaspectratio]{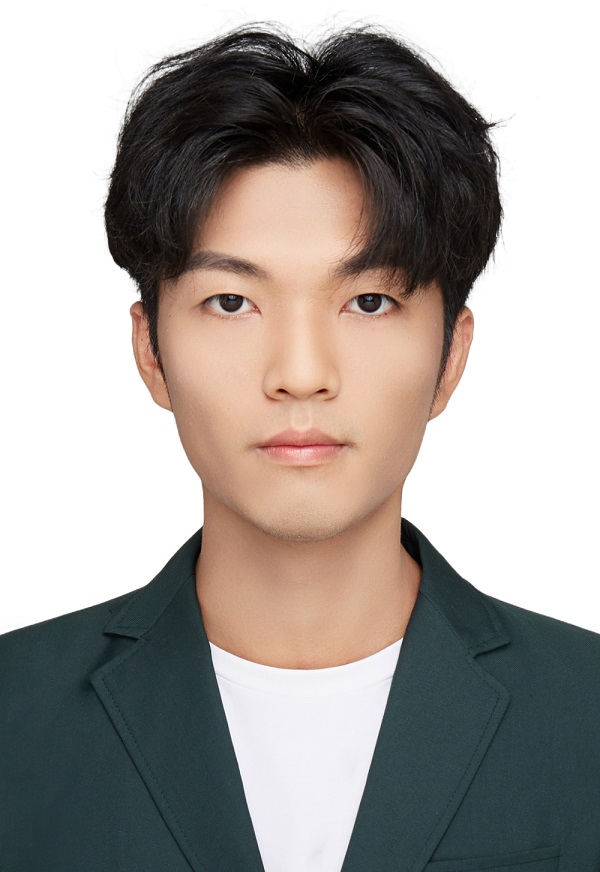}}]{Zixu Wang} is currently pursuing a Ph.D. degree at South China University of Technology, Guangzhou, China. He received the B.E. degree in Automation from Central South University, China, in 2022. His research interests include vision-and-language navigation, test-time adaptation and continual learning.
\end{IEEEbiography}

\begin{IEEEbiography}[{\includegraphics[width=1.0in,height=1.35in,clip,keepaspectratio]{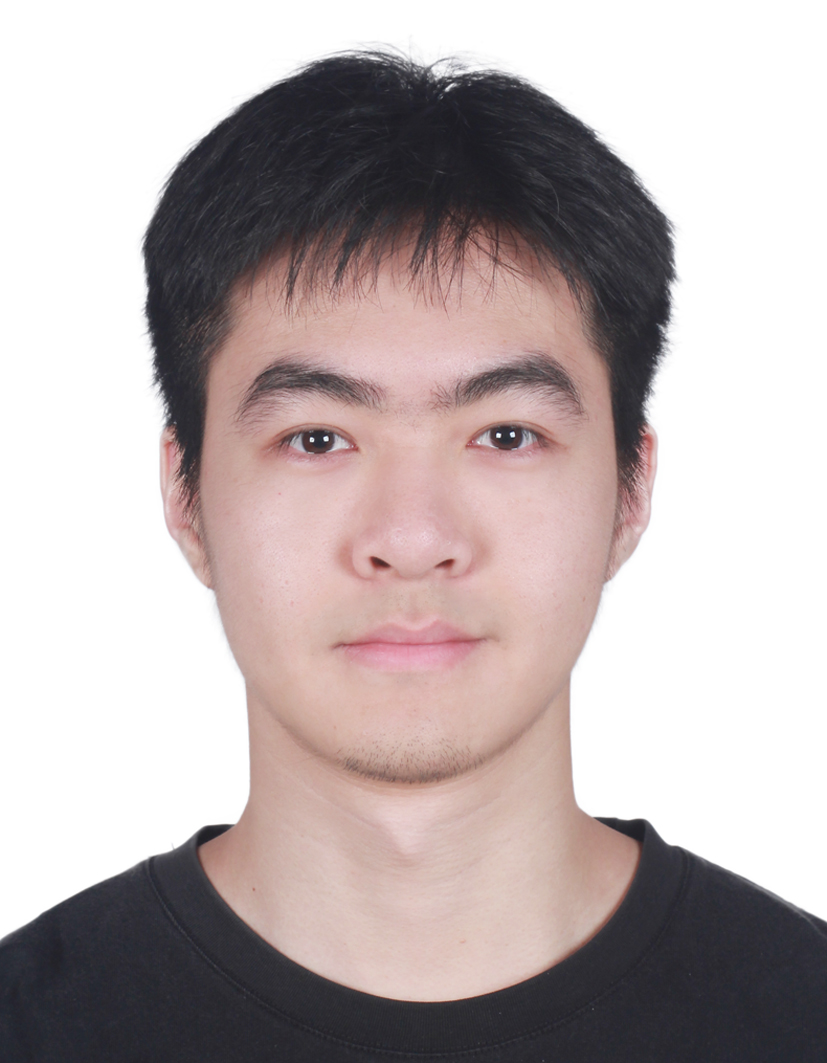}}]{Huaquan Zhang}
is currently pursuing the M.S. degree at South China University of Technology, Guangzhou, China. He received his B.Sc. degree from South China University of Technology, Guangzhou, China, in 2025. His research interests include vision-and-language navigation and test-time adaptation.
\end{IEEEbiography}

\begin{IEEEbiography}[{\includegraphics[width=1in,height=1.25in,clip,keepaspectratio]{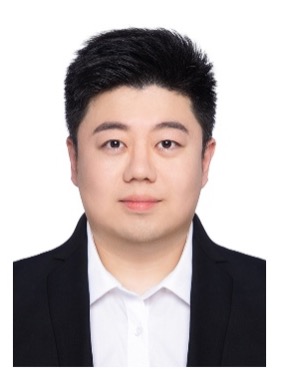}}]{Shoukai Xu} is Postdoctoral Researcher with School of Software Engineering at South China University of Technology. He received his Ph.D. degree in the School of Software Engineering in 2025 from South China University of Technology in Guangzhou, China. His research interests include computer vision and lightweight neural model optimization. He has published papers in top AI venues, including ICML, ICLR, ECCV and IEEE TCSVT.
\end{IEEEbiography}

\begin{IEEEbiography}[{\includegraphics[width=1.0 in,height=1.35in, clip,keepaspectratio]{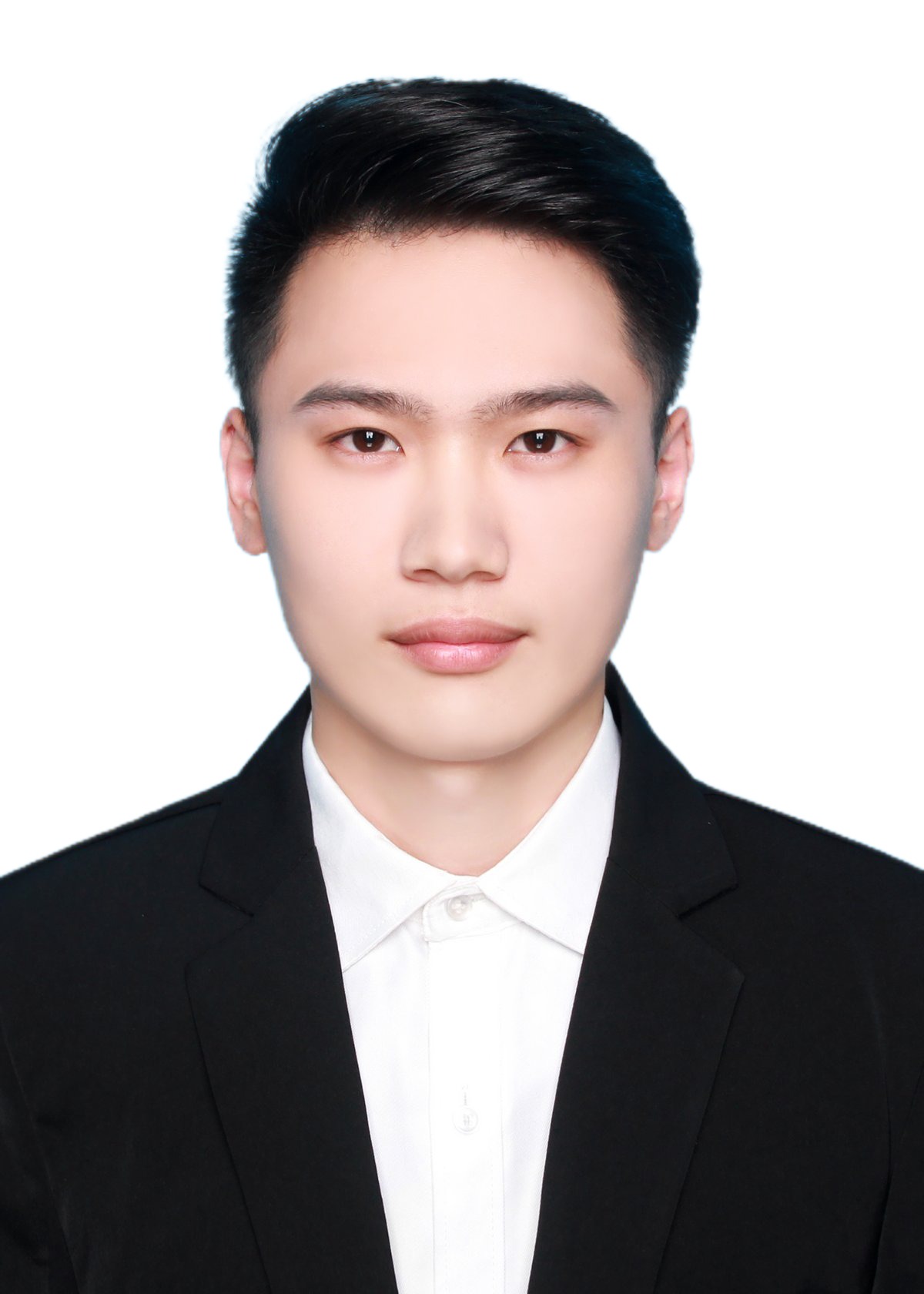}}]{Jinwu Hu} is currently pursuing a Ph.D. degree at South China University of Technology, Guangzhou, China. He received the M.S. degree from Chongqing University of Posts and Telecommunications, Chongqing, China, in 2023. His research interests include LLMs, computer vision, and reinforcement learning. He has published papers in top venues, including ICML, NeurIPS, CVPR, IJCAI, ACM MM, ECCV, IEEE TKDE, IEEE TIP, IEEE TIFS, IEEE TMM, IEEE TCYB, IEEE TMI, etc. He has served as a reviewer for many journals/conferences, including TPAMI, ICML, NeurIPS, CVPR, etc.
\end{IEEEbiography}

\begin{IEEEbiography}[{\includegraphics[width=1in,height=1.25in,clip,keepaspectratio]{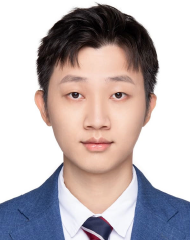}}]{Zhiquan Wen} is recently a Post-doctoral Researcher with Shien-Ming Wu School of Intelligent Engineering at South China University of Technology. He received his Ph.D. degree in the School of Software Engineering in 2024 from South China University of Technology in Guangzhou, China. His research interests include vision-and-language and test-time adaptation. He has published papers in top venues, including NeurIPS, ICLR, ECCV, AAAI, ACL, IEEE TMM, IEEE TCSVT and Neural Networks. He has been invited as a reviewer for top-tier conferences and journals including IEEE TPAMI, IEEE TMM, ICLR, ICML, NeurIPS, CVPR, ICCV, AAAI, etc.
\end{IEEEbiography}
\vspace{-10pt}
\begin{IEEEbiography}[{\includegraphics[width=1.0 in,height=1.35in,clip,keepaspectratio]{{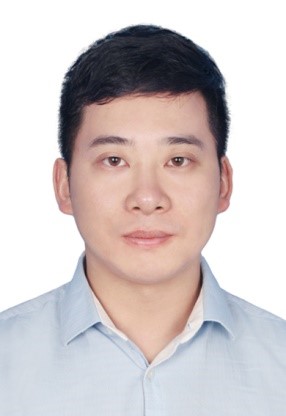}}}]{Mingkui Tan} is currently a Professor with the School of Software Engineering at South China University of Technology. He received his B.Sc. degree in Environmental Science and Engineering and M.Eng. degree in Control Science and Engineering from Hunan University, Changsha, China, in 2006 and 2009, respectively. He obtained his Ph.D. degree in Computer Science from Nanyang Technological University, Singapore, in 2014. From 2014 to 2016, he worked as a Senior Research Associate in the School of Computer Science at the University of Adelaide, Australia, where his research focused on computer vision. He currently serves as an Associate Editor for both IEEE Transactions on Pattern Analysis and Machine Intelligence (TPAMI) and Machine Intelligence Research (MIR). His research interests include machine learning, deep learning, computer vision, sparse analysis, and large-scale optimization.
\end{IEEEbiography}

\vfill

\appendices
\input{supp}

\end{document}

%% file: supp.tex
\twocolumn[
\begin{center}
    \large \bf Supplementary Materials for ``{Asking Less, Adapting Better: Feedback-Efficient Test-Time Adaptation for Vision-and-Language Navigation}''
\end{center}
\vspace{1em}
]

\section{Adaptation Results on R2R}
\label{app:r2r_result}
We further evaluate \methodname{} on the classic R2R benchmark using DUET, GOAT, and ScaleVLN as navigation backbones. As shown in Table~\ref{tab:r2r}, \methodname{} improves overall navigation performance across all three backbones. On the Val Unseen split, it achieves higher navigation success than ATENA while maintaining or improving path efficiency. These results demonstrate the effectiveness of \methodname{} on R2R across different navigation backbones.

\begin{table}[h]
\centering
\renewcommand{\tabcolsep}{1pt}
\caption{Experimental results on R2R dataset.}
\label{tab:r2r}
    \resizebox{\columnwidth}{!}{
        \begin{tabular}{c *{8}{c}}
            \toprule
            \multirow{2}{*}{\textbf{Methods}}
            & \multicolumn{4}{c}{\textbf{Val Seen}}
            & \multicolumn{4}{c}{\textbf{Val Unseen}} \\
            \cmidrule(lr){2-5} \cmidrule(lr){6-9}
            & OSR $\uparrow$ & SR $\uparrow$ & SPL $\uparrow$ & NE $\downarrow$
            & OSR $\uparrow$ & SR $\uparrow$ & SPL $\uparrow$ & NE $\downarrow$ \\
            \midrule
            \multicolumn{1}{l}{DUET~\cite{chen2022think}}
            & 86 & 79 & 73 & 2.28
            & 81 & 72 & 60 & 3.31 \\
            \multicolumn{1}{l}{\hspace{0.5em}\textbullet~FSTTA~\cite{gao2024fastslow}}
            & 83 & 76 & 69 & 2.51
            & 78 & 68 & 57 & 3.47  \\
            \multicolumn{1}{l}{\hspace{0.5em}\textbullet~FeedTTA~\cite{kim2025test}}
            & 85 & 78 & 73 & 2.29
            & 80 & 72 & 61 & 3.26  \\
            \multicolumn{1}{l}{\hspace{0.5em}\textbullet~ATENA~\cite{ko2025active}}
            & 85  & 79  & 74 & 2.28 
            & 79  & 72  & 62  & 3.23  \\
            \rowcolor[HTML]{EEF4FF}
            \multicolumn{1}{l}{\hspace{0.5em}\textbullet~\textbf{\methodname{}}}
            & \textbf{86} & \textbf{80} & \textbf{74} & \textbf{2.20}
            & \textbf{82} & \textbf{74} & \textbf{62} & \textbf{3.01} \\
            
            \midrule
            \multicolumn{1}{l}{GOAT~\cite{Wang2024GOAT}}
            & 89  & 85  & 80  & 1.67  
            & 86  & 78  & 68  & 2.31  \\
            \multicolumn{1}{l}{\hspace{0.5em}\textbullet~FSTTA~\cite{gao2024fastslow}}
            & 88 & 83 & 78 & 1.82
            & 85 & 77 & 66 & 2.45  \\
            \multicolumn{1}{l}{\hspace{0.5em}\textbullet~FeedTTA~\cite{kim2025test}}
            & 89 & 85 & 80 & 1.66 
            & 86 & 78 & 68 & 2.30 \\
            \multicolumn{1}{l}{\hspace{0.5em}\textbullet~ATENA~\cite{ko2025active}}
            & 89  & 85  & 80  & 1.67 
            & 86  & 78  & 69  & 2.27  \\
            \rowcolor[HTML]{EEF4FF}
            \multicolumn{1}{l}{\hspace{0.5em}\textbullet~\textbf{\methodname{}}}
            & \textbf{90} & \textbf{85} & \textbf{80} & \textbf{1.65}
            & \textbf{86} & \textbf{79} & \textbf{69} & \textbf{2.21} \\
            \midrule
            \multicolumn{1}{l}{ScaleVLN~\cite{wang2023scaling}}
            & 87 & 81 & 75 & 2.12
            & 88 & 81 & 70 & 2.09 \\
            \multicolumn{1}{l}{\hspace{0.5em}\textbullet~FSTTA~\cite{gao2024fastslow}}
            & 84 & 78 & 72 & 2.32
            & 86 & 78 & 67 & 2.30  \\
            \multicolumn{1}{l}{\hspace{0.5em}\textbullet~FeedTTA~\cite{kim2025test}}
            & 87 & 82 & 76 & 2.10
            & 87 & 80 & 71 & 2.12  \\
            \multicolumn{1}{l}{\hspace{0.5em}\textbullet~ATENA~\cite{ko2025active}}
            & 86  & 82  & 77  & 2.04 
            & 86  & 79  & 71  & 2.18  \\
            \rowcolor[HTML]{EEF4FF}
            \multicolumn{1}{l}{\hspace{0.5em}\textbullet~\textbf{\methodname{}}}
            & \textbf{88} & \textbf{82} & \textbf{77} & \textbf{1.96}
            & \textbf{88} & \textbf{81} & \textbf{72} & \textbf{2.08} \\
            \bottomrule
        \end{tabular} 
    }
\end{table}

\section{Derivation of Entropy-Gradient Saliency}
\label{app:derivation_saliency}
We derive the closed-form expression for the entropy-gradient saliency $G_t = \left\| \nabla_{\mathbf{z}_t} \hat{H}_t \right\|$ used in the step-level selection criterion (Section~\ref{sec:step_selection}). Let $\mathbf{z}_t = (z_1, z_2, \ldots, z_N)^\top$ denote the action logits at step $t$, where $N = |\mathcal{A}_t|$ is the number of feasible actions. The policy is defined via the softmax function:
\begin{equation}
\pi_j \triangleq \pi_\theta(a_j \mid s_t) = \frac{\exp(z_j)}{\sum_{k=1}^{N} \exp(z_k)}, \quad j = 1, \ldots, N.
\end{equation}
The policy entropy and its normalized form are:
\begin{equation}
H_t = -\sum_{j=1}^{N} \pi_j \log \pi_j, \qquad \hat{H}_t = \frac{H_t}{\log N}.
\end{equation}

\paragraph{Step 1: Softmax Jacobian.}
The partial derivative of the softmax output $\pi_i$ with respect to the logit $z_j$ is:
\begin{equation}
\frac{\partial \pi_i}{\partial z_j} = \pi_i \left( \delta_{ij} - \pi_j \right),
\label{eq:softmax_jacobian}
\end{equation}
where $\delta_{ij}$ is the Kronecker delta.

\paragraph{Step 2: Gradient of the entropy with respect to logits.}
Taking the partial derivative of $H_t$ with respect to $z_j$:
\begin{align}
\frac{\partial H_t}{\partial z_j}
&= -\sum_{i=1}^{N} \frac{\partial}{\partial z_j} \left[ \pi_i \log \pi_i \right] \notag \\
&= -\sum_{i=1}^{N} \left[ \frac{\partial \pi_i}{\partial z_j} \log \pi_i + \pi_i \cdot \frac{1}{\pi_i} \frac{\partial \pi_i}{\partial z_j} \right] \notag \\
&= -\sum_{i=1}^{N} \frac{\partial \pi_i}{\partial z_j} \left( \log \pi_i + 1 \right).
\label{eq:dH_dz_expanded}
\end{align}
Substituting the softmax Jacobian from Eq.~\eqref{eq:softmax_jacobian}:
\begin{align}
\frac{\partial H_t}{\partial z_j}
&= -\sum_{i=1}^{N} \pi_i \left( \delta_{ij} - \pi_j \right) \left( \log \pi_i + 1 \right) \notag \\
&= -\pi_j \left( \log \pi_j + 1 \right) + \pi_j \sum_{i=1}^{N} \pi_i \left( \log \pi_i + 1 \right).
\label{eq:dH_dz_split}
\end{align}
Note that $\sum_{i=1}^{N} \pi_i \log \pi_i = -H_t$ and $\sum_{i=1}^{N} \pi_i = 1$. Therefore:
\begin{align}
\frac{\partial H_t}{\partial z_j}
&= -\pi_j \left( \log \pi_j + 1 \right) + \pi_j \left( -H_t + 1 \right) \notag \\
&= -\pi_j \log \pi_j - \pi_j + \pi_j - \pi_j H_t \notag \\
&= -\pi_j \left( \log \pi_j + H_t \right).
\label{eq:dH_dz_final}
\end{align}

\paragraph{Step 3: Gradient of the normalized entropy.}
Since $\hat{H}_t = H_t / \log |\mathcal{A}_t|$ and $|\mathcal{A}_t|$ is treated as a constant at step $t$:
\begin{equation}
\frac{\partial \hat{H}_t}{\partial z_j} = \frac{1}{\log |\mathcal{A}_t|} \cdot \frac{\partial H_t}{\partial z_j} = -\frac{\pi_\theta(a_j \mid s_t) \left( \log \pi_\theta(a_j \mid s_t) + H_t \right)}{\log |\mathcal{A}_t|}.
\label{eq:dHhat_dz}
\end{equation}

\paragraph{Step 4: Entropy-gradient saliency.}
The saliency score is defined as the $\ell_2$-norm of the gradient vector. Substituting Eq.~\eqref{eq:dHhat_dz}:
\begin{align}
G_t
&= \left\| \nabla_{\mathbf{z}_t} \hat{H}_t \right\|
= \sqrt{ \sum_{a_j \in \mathcal{A}_t} \left( \frac{\partial \hat{H}_t}{\partial z_j} \right)^2 } \notag \\
&= \frac{1}{\log |\mathcal{A}_t|} \sqrt{ \sum_{a_j \in \mathcal{A}_t} \pi_\theta\left(a_j \mid s_t\right)^2 \left( \log \pi_\theta\left(a_j \mid s_t\right) + H_t \right)^2 }.
\label{eq:saliency_final}
\end{align}

This expression can be computed in closed form from the softmax probabilities and the entropy, requiring no backpropagation through the model.

\begin{figure*}[t]
    \centering
    \caption{Human annotation interface for binary action feedback. Annotators assess action usefulness using the task instruction, trajectory video, and panoramic observations before and after the queried action.}
    \includegraphics[width=\textwidth]{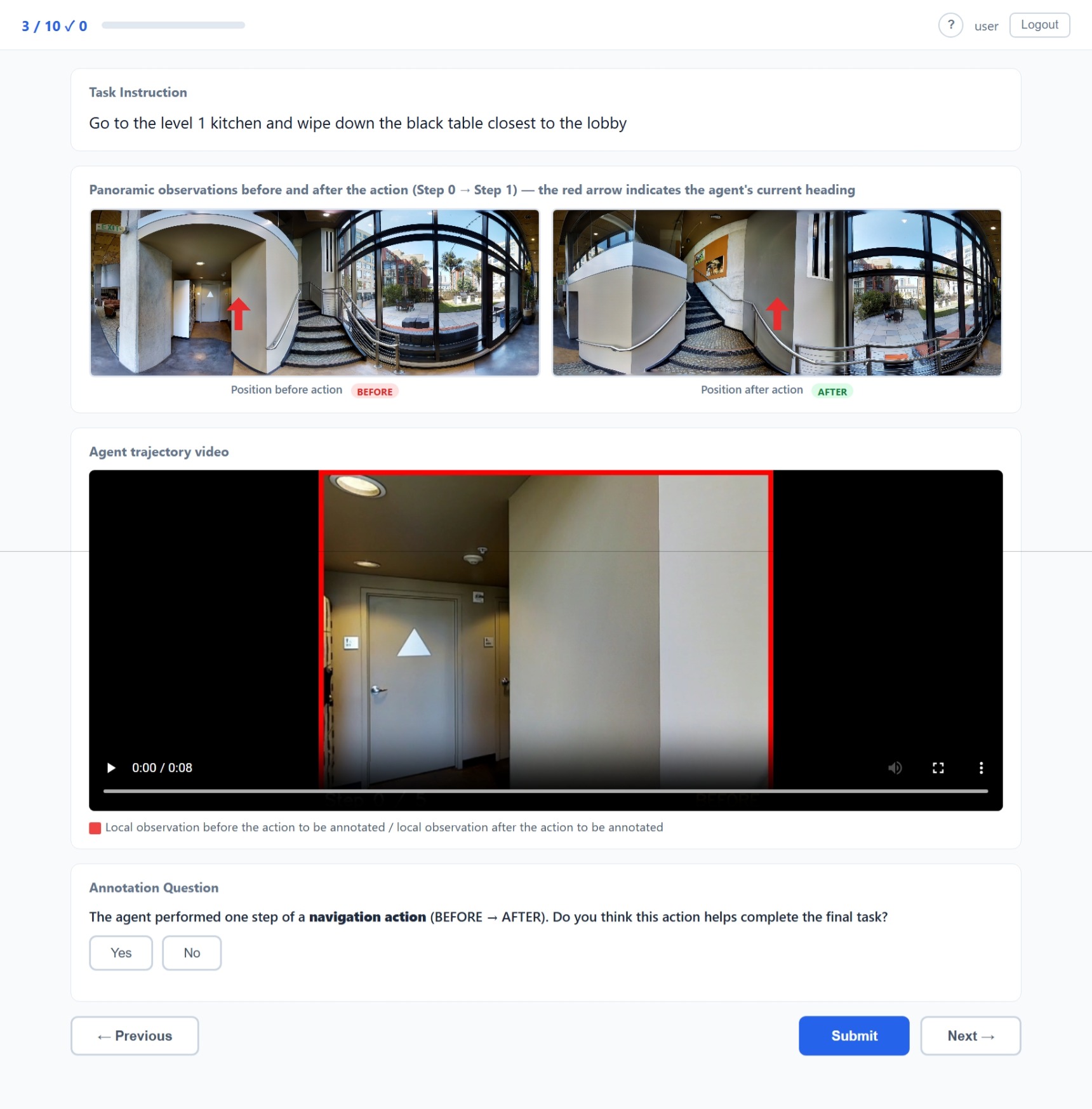}
    \label{fig:label_page}
\end{figure*}

\vfill